%% file: main.tex
\pdfoutput=1
\documentclass[11pt]{article}

\usepackage[]{ACL2023}
 
\usepackage{microtype}
\usepackage{hyperref}
\usepackage{nameref}

\usepackage{times}
\usepackage{latexsym}
\usepackage{graphicx}
\usepackage{algorithm, algorithmic}
\usepackage{amsmath}
\usepackage{booktabs}
\usepackage{tabularx}
\usepackage{colortbl}
\usepackage{hhline}
\usepackage{amssymb} 
\usepackage{array} % 导入array宏包以使用m{width}等功能
\usepackage{multirow} % For multirow cells
\usepackage{enumitem}
\usepackage[T1]{fontenc}
\usepackage[most]{tcolorbox}
\newtcolorbox{prompt}[1]{
    enhanced,
    drop shadow=black!5!white,
    left=4mm,
    right=4mm,
    top=2mm,
    bottom=2mm,
    boxsep=0mm,
    rounded corners,
    title=#1,
    fontupper=\footnotesize\linespread{0.9}\fontfamily{lmr}\selectfont,
    }
\usepackage[utf8]{inputenc}

\usepackage{inconsolata}

\title{BodyCam-VQA: Enhanced Body-Worn Camera Video Captioning via Multimodal Reasoning and Probe Question Generation}

\author{
\textbf{Karish Gupta}\textsuperscript{$\clubsuit$} \quad
\textbf{Matthew Alex}\textsuperscript{$\clubsuit$} \quad
\textbf{Alex Li}\textsuperscript{$\clubsuit$} \\
\textbf{Yang Wu}\textsuperscript{$\clubsuit$} \quad
\textbf{Yun-Wei Chu}\textsuperscript{$\heartsuit$} \quad
\textbf{Kashif Munir}\textsuperscript{$\heartsuit$} \\
\textbf{Xiaotian Zhou}\textsuperscript{$\heartsuit$} \quad
\textbf{Zhengping Ji}\textsuperscript{$\heartsuit$} \quad
\textbf{Xiaozhong Liu}\textsuperscript{$\clubsuit$}\thanks{\, Corresponding author.} \\
\textsuperscript{$\clubsuit$}Worcester Polytechnic Institute, Worcester, MA, USA \\
\textsuperscript{$\heartsuit$}Axon, Scottsdale, AZ, USA \\
\texttt{\{kagupta, malex, ajli, ywu19, xliu14\}@wpi.edu} \\
\texttt{\{ychu, kmunir, xzhou, zji\}@axon.com}
}

\begin{document}
\maketitle

\input{sections/0_abstract}

\input{sections/1_introduction}

\input{sections/2_related_work}

\input{sections/3_methodology}

\input{sections/4_experiment}

\input{sections/5_conclusion}

\input{sections/6_limitations}

\input{sections/7_ethics_statement}

\bibliography{anthology,custom}
\bibliographystyle{acl_natbib}

\nocite{}
\clearpage 
\input{sections/X_appendix}

\end{document}

%% file: sections/0_abstract.tex
\begin{abstract}

Police body-worn camera (BWC) footage has emerged as a critical aspect of law enforcement that ensures legal transparency, officer accountability, and the protection of civil rights. However, effectively processing this data remains a significant challenge due to its multimodal video format. BWC videos, in many cases, comprise chaotic scenes with low visual quality, rapid movement/interactions, and high-noise audio that make visual understanding a challenge for even SOTA multimodal models. Current Vision-Language Models (VLMs) frequently overlook critical forensic details, such as the presence of valuable evidence or the latent nuances of suspect-officer interactions, which are vital for fair legal outcomes and civilian/officer safety. To address these limitations, we propose an Adaptive Visual Question Answering (VQA) framework engineered for high-stakes law enforcement. Our framework employs a structured reasoning approach to extract fine-grained visual evidence that traditional captioning systems fail to capture. We experiment with multiple question generation models, including foundation models and fine-tuned open-weight models, to observe performance variation among question generation model implementations. Our results demonstrate that this VQA-driven architecture provides a more reliable, objective, and detailed record of enforcement events, ultimately serving as a powerful tool to protect both law enforcement officers and the public through AI-assisted forensic clarity.\footnote{https://github.com/Karish-Gupta/BodyCam-VQA}

\end{abstract}

%% file: sections/1_introduction.tex
\section{Introduction}

The integration of police body-worn cameras (BWCs) into modern law enforcement has allowed a large amount of transparency into officer activity; however, many overlook the massive intelligence asset that BWC footage has become for the officers themselves~\citep{voigt2017language,lum2020body,camp2024leveraging}. By providing an objective view of high-stakes law enforcement encounters, these devices can be used to gather a great deal of evidence about law enforcement encounters that may be overlooked without video evidence~\citep{white2014police}. A 2022 study looking into the impact of body-worn cameras on police report writing showed that participants who did not review the BWC footage of police encounters were almost four times more likely to make at least one factual error in their written report~\cite{Boivin2022}. However, manual report writing is an exceptionally tedious task and can be difficult to implement at large scale when there is a great deal of video footage~\citep{ferguson2025generative,watts2026ai}. Thus, with the growing domain of multi-modal large language models (MLLMs) and vision language models (VLMs), we see potential in building captioning pipelines for complex police BWC footage. Similar work has been done with the OpenBWC framework~\cite{srbinovska2025aidrivenpolicinginterdisciplinaryknowledge}, which proposes a multi-modal system to analyze interactions and behaviors found in BWC footage. This model, potentially valuable for law enforcement training, does not directly extract details and information from a given scene. We aim to build a system that can understand police interactions, point out valuable evidence (objects, people, etc.), and describe in detail the setting and location of a given video.
% TOY EXAMPLE
\begin{table*}[t] % The asterisk * makes it span two columns
\centering
\small
\begin{tabularx}{\textwidth}{@{}l X@{}}
\toprule
\textbf{Pipeline Stage} & \textbf{Descriptions and Generated Outputs} \\ \midrule
\textbf{Input Video} & 60s BWC footage: Video depicts an officer chasing a suspect trying to flee, deploying a Taser, then wrestling and detaining the person against a parked car\\ \addlinespace
\textbf{Initial VLM Summary} & Officer pursues a civilian in a white t-shirt at 01:55:29. Taser deployed at 01:55:45. Physical struggle occurs against a parked vehicle. \\ \addlinespace
\textbf{Generated VQA} & 
\textbf{Q:} What is the duration of the grapple? \textbf{A:} Approximately 20 seconds. \\
& \textbf{Q:} Are additional officers present? \textbf{A:} Yes, audio cues and brief visual frames imply multiple officers. \\ \addlinespace
\textbf{Final QA Caption} & At 01:55:15, a BWC reveals a dark street with flashing police cruisers. Following a foot pursuit (01:55:29) and Taser deployment (01:55:45), the officer engages in an \textbf{intense 20-second struggle}. Notably, overlapping voices and visual cues confirm the \textbf{presence of additional officers}, concluding with the subject pinned against a parked SUV. \\ \bottomrule
\end{tabularx}
\caption{Qualitative BodyCam-VQA Walkthrough}
\label{tab:toy_example}
\end{table*}

BWC videos frequently contain chaotic visual content and noisy audio conditions, making most types of automated understanding much more challenging~\citep{grauman2022ego4d}. Our work provides a framework for enhanced captioning of these videos, leading us to believe that a visual question-answering (VQA) system, alongside a strong VLM, could be a viable option for detailed and accurate automated video captioning.

Prior work on VQA has primarily focused on well-structured domains such as medical and surgical videos. Representative examples include LLaVA-Surg~\cite{llavasurg2024} and Surgical-VQA~\cite{surgicalvqa2022}, which demonstrate the effectiveness of question-driven supervision for extracting fine-grained visual semantics. The LLaVA-Surg framework developed a two-stage question-answer generation pipeline: the framework first extracts discrete, structured factual information from surgical lecture videos and then uses pairs of structured information to generate intelligent questions. This approach greatly reduces the major risk of generating hallucinations in the questions.

Our goal is to utilize visual question-answering in tandem with powerful multimodal models to enhance the quality of BWC video captioning. A qualitative example of how this change can be observed in the generated captions is laid out in Table~\ref{tab:toy_example}, where we can observe valuable details incorporated into the original caption as a direct result of intelligent and specific VQA. The addition of the questions pushed the model to look for more individuals in the scene, which changed the narrative from a single-officer scene to one of multi-officer involvement. This directly increases the contextual density of the generated caption.

In summary, our primary contributions to this work are as follows: 

\begin{itemize}[nosep]
    \item We construct a new Body-Worn Camera (BWC) video dataset comprising 359 annotated 60-second videos with aligned transcripts to support multi-modal reasoning and analysis.
    \item We propose BodyCam-VQA, an adaptive Visual Question Answering framework tailored for body-worn camera footage. The architecture is designed to handle chaotic scenes, low visual quality, rapid interactions, and noisy audio by employing structured reasoning to extract fine-grained forensic evidence that conventional captioning and existing Vision-Language Models often overlook.
    \item We empirically validate the effectiveness of BodyCam-VQA through comprehensive experiments and human evaluation, demonstrating improvements in factual accuracy, answer completeness, and visual grounding.
\end{itemize}

% \begin{figure}[t]
% \centering 
% \includegraphics[width=\linewidth]{figures/toy_example.pdf}
% \caption{Toy example.} 
% \label{fig:intro}
% \end{figure}

%% file: sections/2_related_work.tex
\section{Related Work}

\subsection{Video Understanding}
Video understanding in the past few years has shifted towards structured multimodal reasoning~\cite{fu2026novaplan}. Vision-Language Models (VLMs) have had the biggest impact, using methods such as visual instruction tuning~\cite{liu2024visualinstruction}, self-distillation~\cite{naeem2023silc}, and reinforcement learning~\cite{gdpo2026, guo2025mmrl,wu2026excuse} to help improve the alignment of these VLMs from recognition to decision making. In more complex settings, Chain-of-Thought(CoT) reasoning was incorporated into the multimodal domain, improving performance~\cite{zhang2025visualchains, park-etal-2025-making}. Despite many improvements in the field, many blockers are still in place. These mainly focus on temporal hallucination~\cite{fu2024videomme}, where VLMs fabricate events in longer-form videos. This is because many of these models sample a few frames from thousands, which removes important context required for accurate reasoning. Models also struggle to distinguish between similar actions (e.g., handshakes vs fist-bumps), leading to major problems in complex reasoning, where these mistakes can invalidate a long reasoning chain. 

\subsection{Visual Question Answering}
Visual Question Answering (VQA) has emerged as a powerful paradigm for extracting fine-grained, targeted information that general-purpose captioning often overlooks~\cite{gupta2017survey, huang2021survey}, with structured questions shown to elicit more detailed, factually grounded responses than holistic descriptions. This has proven effective in well-structured domains such as radiology~\cite{lau2018dataset}, pathology~\cite{he2020pathvqa}, long-form video~\cite{yu2019activitynet}, and legal consultation~\cite{wu2024knowledge, yao2025elevating}. LLaVA-Surg~\cite{llavasurg2024} and Surgical-VQA~\cite{surgicalvqa2022} further show that question-driven supervision extracts fine-grained surgical details missed by direct captioning. However, these works remain confined to controlled, well-lit, single-task environments, leaving open whether this paradigm extends to chaotic, multi-participant settings such as BWC footage.

\subsection{VLMs in Public Safety}
In the domain of public safety, two papers investigate similar topics. The first paper introduces VIVID~\cite{Gonzalez2025VIVID}. This model focuses on identifying high-motion keyframes through optical flow, which are most likely to contain violent actions. It then converts the actions into text tokens and compares the generated text to the formal definitions of crimes to determine if they reach the standard.
The second paper is an evaluation paper of VLMs on Surveillance Video~\cite{Benschop2025Evaluation}, testing the power of smaller models (8B or fewer parameters) on anomaly detection. In their model, the VLM generates a description, and a separate model scores them based on whether they contain an anomaly label such as "theft". The paper highlighted that the models were good at detecting events when they were "clear" (a fight in the center of the frame), but not as good at detecting events that were noisy or partially obstructed.

%% file: sections/3_methodology.tex
\section{Methodology}

\subsection{Problem Definition}
The BodyCam-VQA pipeline is designed to transform multimodal inputs into dense, investigative captions. Given a BWC video $V$ and its associated transcript $T$, our goal is to synthesize a VQA-enhanced caption $C$. This caption represents a comprehensive analysis of the scene, capturing scene-level context, the individuals involved, and critical environmental details. 

\subsection{BodyCam-VQA Mechanism}
% Give brief explanation of worklfow and say we will introduce them in next sections

The framework $\mathcal{F}$ consists of four primary stages:
\begin{itemize}[nosep]
    \item Multimodal Summarization ($S$)
    \item Structured Detail Extraction ($D$)
    \item Targeted Question Generation ($Q$)
    \item Visual Question Answering and Synthesis ($C$)
\end{itemize}
We will introduce each part of the workflow in the following sections.

% PIPELINE DIAGRAM
\begin{figure}[h]
    \centering
    \includegraphics[width=0.75\linewidth]{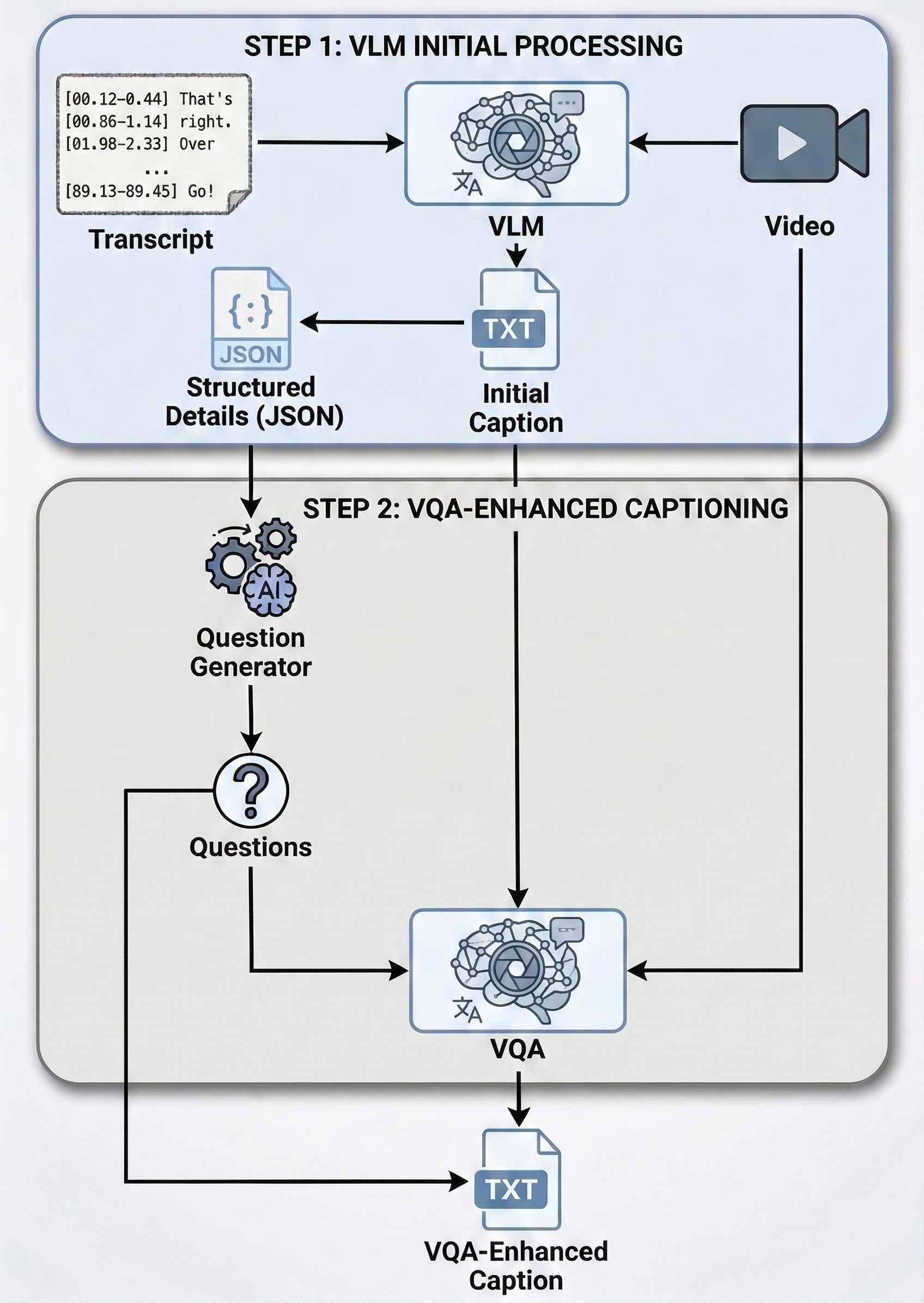}
    \caption{Overview of the BodyCam-VQA Framework and its Two Main Steps. Initial Caption Generation Creates a Baseline VLM Summarization. Final VQA-Enhanced Caption Utilizes VQA Driven by Targeted Question Generation.}
    \label{fig:Pipeline}
\end{figure}

\subsubsection{VLM Summarization}
Preliminary experiments indicated that audio transcripts $T$ often lack the visual granularity necessary for high-quality caption generation. All visual details (e.g., physical interactions, objects in the scene, participant descriptions) are overlooked. To bridge this gap, we utilize a VLM to generate an initial visual summary $S = f_{VLM}(V, T)$. This summary serves a dual purpose: it provides the primary visual context for subsequent stages and acts as a baseline to evaluate the performance gain of our multi-pass pipeline over standard single-pass VLM captioning.

\subsubsection{Structured Details}
To ensure focused investigative questions, we adapt the LLaVA-Surg~\cite{llavasurg2024} framework. Prior work in PDE-based Bayesian hierarchical modeling~\cite{cen2025pde} shows that decomposing complex processes mitigates observational noise. Inspired by this, we extract structured details $D$ in JSON format as observed in Equation~\ref{eq:structured_details}
\begin{equation}
    \mathcal{D} = \{D_{\text{scene}}, D_{\text{entity}}, D_{\text{action}}, D_{\text{semantic}}\}
\label{eq:structured_details}
\end{equation}

\begin{figure}[th]
    \centering
    \includegraphics[width=0.75\linewidth]{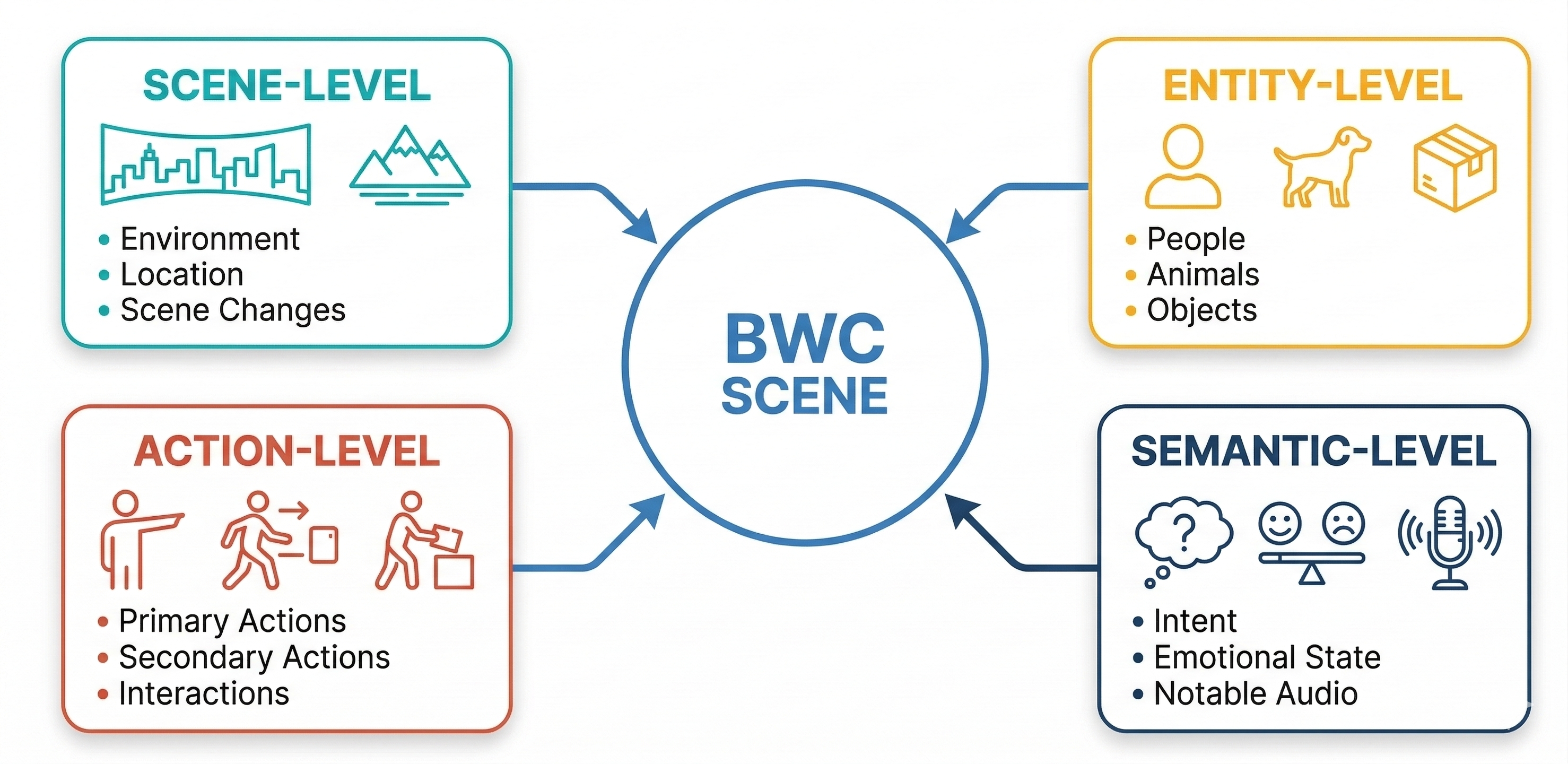}
    \caption{Structured Details Schema. Comprised of Scene, Entity, Action, and Semantic Level Details.}
    \label{fig:Structured_details}
\end{figure}

This process compresses raw multimodal information into a discrete set of fine-grained categories (scene-level, entity-level, action-level, and semantic-level), as illustrated in Figure~\ref{fig:Structured_details}. By grounding question generation in these structured details, we enforce factual consistency and mitigate the risk of model hallucination during the question generation phase.

\subsubsection{Question Generation Models}
The efficacy of the VQA enhanced caption $C$ is inherently based on the quality of the generated questions $Q$. To optimize this, we formalize question generation as:
\begin{equation}
Q = G(D)
\label{eq:structured_details}
\end{equation}
Where $G$ is a generative model. We investigated various architectures for $G$, ranging from large-scale foundation models to specialized, fine-tuned models.

% STRUCTURED DETAILS
\subsubsection{Training Data with Chain of Thought}
Preliminary evaluations indicated that smaller language models struggled to consistently generate investigative questions capable of extracting nuanced details from BWC footage. In contrast, larger proprietary models demonstrated significant proficiency in this domain. To bridge this performance gap, we implemented multiple fine-tuning approaches, utilizing a generated dataset of high-quality questions with chain-of-thought reasoning (to encourage smaller models to follow a similar thought process). This training data of curated high quality questions, paired with latent reasoning, was generated with a proprietary model with strong reasoning and CoT capabilities.

\subsubsection{Standard Fine-tuned Model}
We conducted supervised fine-tuning on open-weight models to evaluate the efficacy of training smaller models on reasoning-dense data. The training set consisted of 288 examples of input structured details, CoT tokens, and high-quality questions. This allows for a direct assessment of how distilled reasoning affects the zero-shot investigative performance of smaller open-weight models.

\subsubsection{GRPO Fine-tuned Model}
To refine the precision and relevance of our investigative questions, we employ Group Relative Policy Optimization (GRPO) \cite{shao2024deepseekmath}.  Unlike SFT, GRPO optimizes the model policy by evaluating a group of sampled outputs against their collective mean, facilitating stable reinforcement learning. By utilizing a critic-based reward, this method enables the model to iteratively improve question quality and investigative depth, surpassing the static performance of SFT.

We define the total reward $R$ for generated questions as a multi-objective function. For a given input $D$, the model generates a set of $G$ outputs $\{o_1, o_2, ..., o_G\}$. The reward for each output $o_i$ is calculated as:
\begin{equation}
\mathcal{R}(o_i) = \alpha \cdot r_{format}(o_i) + \beta \cdot r_{judge}(o_i)
\label{eq:structured_details}
\end{equation}

Where:

\begin{itemize}[nosep]
    \item $r_{format}$ (Rule-based Reward): This component encourages the structural integrity of the generated output. It assigns high values for the correct use of <think> (Chain-of-Thought) and <question> tags. Additionally, it applies a length penalty on short questions and a keyword bonus to prioritize interrogative words (e.g., how, describe, identify). 
    \item $r_{judge}$ (Model-based Reward): To capture semantic nuance, we employ a large foundation model as an automated judge. The judge evaluates each generated question $q \in o_i$ on a binary scale $\{0, 1\}$ based on its relevance to the specific video context $S$ and investigative quality. 
    \item $\alpha, \beta$: These represent coefficients used to balance structural compliance and question quality.
\end{itemize}

\subsubsection{VQA and Caption Synthesis}

In the final stage, the pipeline performs a second VLM pass on the video. The questions are passed to the VLM for visual question-answering, yielding a set of answers $A$. Finally, the original VLM summary $S$, the questions $Q$, and the answers $A$ are ingested by a LLM to produce a final synthesized caption $C$: 
\begin{equation}
C = \text{LLM}(S, Q, A)
\label{eq:structured_details}
\end{equation}

This iterative refinement allows the model to include interactions and objects that may typically be overlooked in a single summarization pass.  

The structured details are passed to the VLM once again, but this time for visual question-answering. This acts like the second VLM pass on the video. However, instead of focusing on broad details, the model is encouraged to observe specific interactions, objects in the scene, etc. Finally, the questions, answers, and original VLM summary are all passed to an LLM for VQA-enhanced caption synthesis. This brings us to the end of the BodyCam-VQA pipeline.

% \begin{figure*}[t]
%   \centering
%   \includegraphics[width=1.0\linewidth]
%   {figures/framework.jpeg}
%   \caption{Illustration of our proposed framework.}
%   \label{framework}
% \end{figure*}

%% file: sections/4_experiment.tex
\section{Source Dataset and Video Preprocessing}
We utilized the Chicago Civilian Office of Police Accountability (COPA) dataset, an open-source collection of incidents involving police, which includes body-worn camera footage with audio. From this dataset, we sampled 40 full-length videos covering a diverse range of police incidents and visual contexts.

% DATASET DIAGRAM
\begin{figure}
    \centering
    \includegraphics[width=0.5\linewidth]{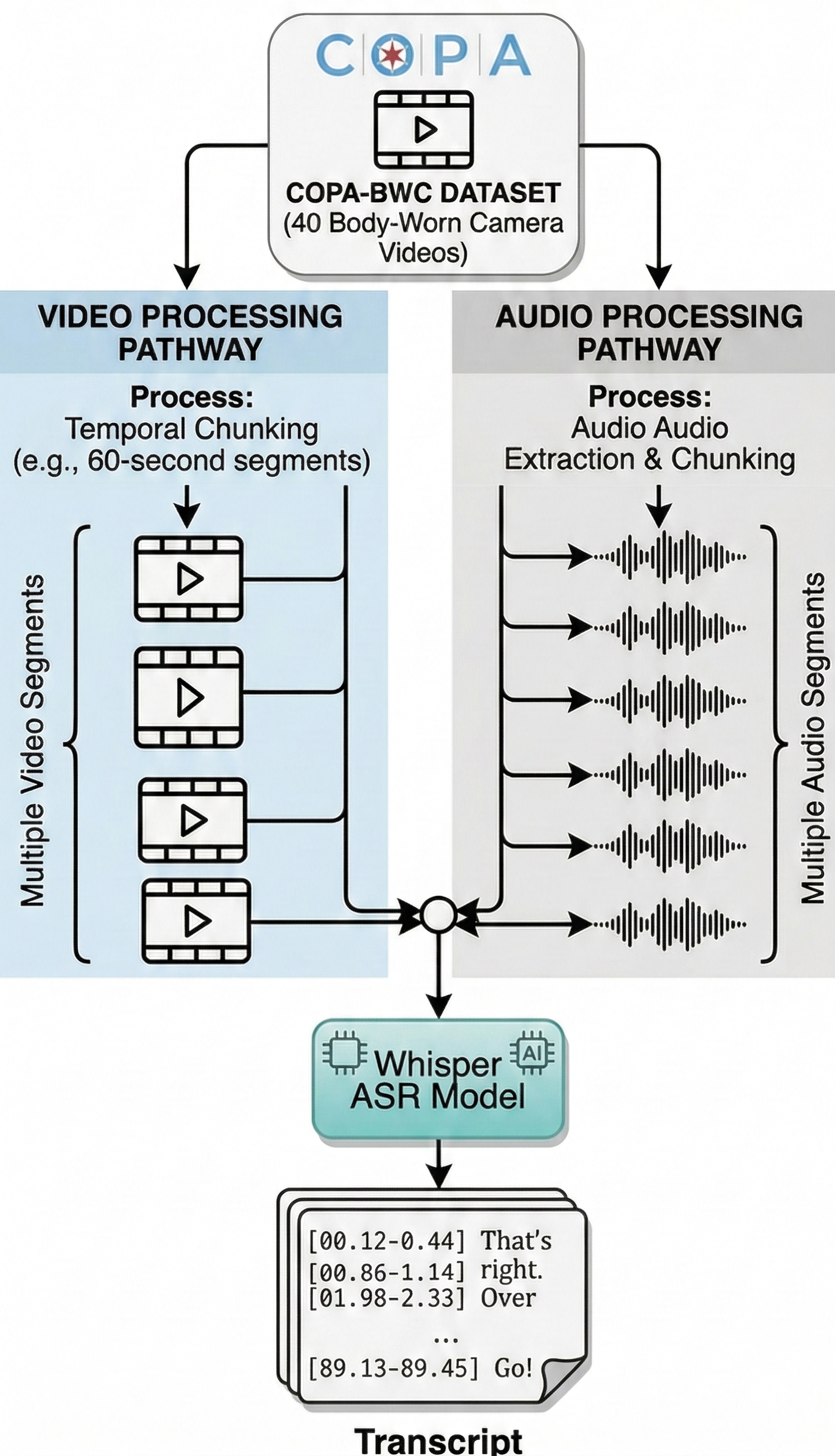}
    \caption{Dataset Preprocessing Pipeline. Ingests Full-Length BWC Videos and Generates 60-Second Video Segment and Transcript Pairs.}
    \label{fig:Dataset}
\end{figure}

To reduce computational overhead and keep our VQA pipeline manageable with limited resources, each video is segmented into 60-second clips, giving us a total dataset size of 359 clips. These clips were chosen from high-action videos from the COPA dataset, and frames were sampled at a rate of 1 FPS. Based on previous works~\cite{vlmsurvey2025, contextlength2024}, current VLMs generally perform best with short videos and relatively simple, visual questions. Since VLMs have to extract a set number of frames from a video, the frame rate can become quite limited. Thus, this clipping strategy helps retain sufficient temporal information for reasoning over actions and events, and allows for traceable inference for large-scale VLMs. In our pipeline, each resulting clip is treated as an independent training instance for downstream processing.

For each 1-minute video segment, we generate high-quality textual transcripts using WhisperX~\cite{whisperx2023}, a state-of-the-art automatic speech recognition (ASR) model. The quality of these captions was checked through manual human checking of each. WhisperX is selected for its highly accurate transcription capabilities. The transcription pipeline produces labeled and time-aligned transcripts per video clip as observed in Figure~\ref{fig:Dataset}.
 
 \section{Experiment}
\subsection{Experimental Settings}
% model, temperature, ss backbone, what xx
Gemini 2.5 Flash is used for initial VLM summarization, VQA, and final caption synthesis in our framework. We selected this proprietary model due to its robust performance and significantly lower hallucination rates in the context of chaotic BWC footage compared to open-weight alternatives. To ensure reproducibility across experiments, the decoding temperature is set to 0.1. 

To establish an open-source performance VLM summary baseline, we also implement Qwen3-VL, a highly capable open-weight VLM currently available. We specifically utilize Qwen3-VL to generate single-pass summaries, allowing us to quantify the performance gains achieved by our proposed multi-pass VQA framework.

For the question generation task, we evaluated both large-scale foundation models and specialized fine-tuned architectures. The foundation models include Gemini 2.5 Flash and DeepSeek V3.2. To improve question quality, we implemented and compared three fine-tuned variants on Qwen3 \footnote{https://huggingface.co/BodyCam-VQA}, a specialized reasoning model optimized for complex text generation and chain-of-thought (CoT) logic:

\begin{itemize}
    \item Qwen3-4B-Thinking-2507 (SFT)
    \item Qwen3-30B-A3B-Thinking-2507 (SFT)
    \item Qwen3-4B-Thinking-2507 (GRPO)
\end{itemize}

All fine-tuned models were trained and deployed using Low-Rank Adaptation (LoRA) to maximize parameter efficiency. For inference, we employed 4-bit quantization to maintain high throughput. The generation parameters were standardized with a temperature of 0.1 and a top-$p$ value of 0.9. All computational experiments were conducted on a single NVIDIA H200 GPU.

\subsection{Baselines}
To evaluate the performance of the proposed architecture, we compare our results against four distinct baselines categorized into two groups: Direct VLM Summarization and Pipeline-based VQA.

Direct VLM Summaries: We employ Gemini 2.5 Flash and Qwen3-VL-32B-Instruct to generate single-pass captions. These models serve as high-performing proprietary and open-source benchmarks, respectively. By evaluating these initial summaries, we establish a performance floor to quantify the incremental value provided by our multi-stage VQA process.

VQA-Integrated Pipelines: We implement two baseline versions of the BodyCam-VQA architecture. In these configurations, the core architecture remains constant, but the question generation (QG) module is powered by either Gemini 2.5 Flash or DeepSeek V3.2. These baselines demonstrate the upper-bound efficacy of the proposed framework when paired with state-of-the-art foundation models and provide a rigorous benchmark for our fine-tuned QG models.

\subsection{Tasks and Metrics}
In order to properly evaluate generated captions for these videos, we put together a human-evaluated ground truth dataset that includes important details, visual enrichment details, and extra auxiliary details from the videos. Important details contain all major events that occurred during the video, visual enrichment details include all important visual details from the video, and auxiliary details include any extra details we added. Our evaluation is based on our ground truth dataset.

Generated captions are evaluated using the following three metrics:
\begin{itemize}[nosep]
    \item "Factual Accuracy": 0 - 1,
    \item "Completeness": 0 - 1,
    \item "Visual Enrichment": 0 - 1,
\end{itemize}

Factual Accuracy: Measures whether the statements in the caption are based on ground truth. If the general idea is mentioned in the ground truth (any section), it is considered true. If extra details are added that contradict the ground truth, the score is penalized. Each sentence must be checked against the ground truth to verify accuracy, and the final score is calculated by dividing the number of statements grounded in truth by the total number of statements. Since we cannot incorporate all details that the model may potentially extract into the ground truth dataset, we used human evaluation to calculate the factual accuracy for each model.

Completeness: Measures whether the caption captures all relevant and important events. The model first retrieves the list of events from the important details section of the ground truth. The model then checks whether each event from the list is present in the generated caption. The final score is the number of included events divided by the total number of important details events.

Visual Enrichment: Measures the proportion of specific visual details captured in the caption. The model first retrieves the list of items from the visual enrichment details section of the ground truth and checks whether each specific visual detail from that list is described in the generated caption. The final score is the number of included visual details divided by the total number of visual enrichment items.

Since there is potential for the completeness and visual enrichment calculations to differ based on the judge, we utilized two judges (Gemini 2.5 Flash and Deepseek V3.2) and took the averages of their judgment results.

\subsection{Main Results}

% MAIN RESULTS TABLE
\begin{table*}[t]
\centering
\small
\begin{tabular}{@{}lcccccc@{}}
\toprule
\textbf{Metric} & \textbf{\begin{tabular}[c]{@{}c@{}}Visual Sum.\\ (Qwen3)\end{tabular}} & \textbf{\begin{tabular}[c]{@{}c@{}}Visual Sum.\\ (Gemini)\end{tabular}} & \textbf{\begin{tabular}[c]{@{}c@{}}Gemini\\ QA\end{tabular}} & \textbf{\begin{tabular}[c]{@{}c@{}}Deepseek\\ QA\end{tabular}} & \textbf{\begin{tabular}[c]{@{}c@{}}Qwen3 30B\\ SFT QA (ours)\end{tabular}} & \textbf{\begin{tabular}[c]{@{}c@{}}Qwen3 4B\\ GRPO QA (ours)\end{tabular}} \\ \midrule
Human Factual Acc. & 0.86 & 0.86 & 0.89 & 0.90 & 0.86 & \textbf{0.91} \\ \midrule
\textit{Gemini as Judge} & & & & & & \\
Completeness & 0.27 & 0.51 & 0.64 & 0.59 & \textbf{0.65} & 0.56 \\
Visual Enrichment & 0.60 & 0.59 & 0.65 & \textbf{0.68} & 0.62 & 0.66 \\ \midrule
\textit{Deepseek as Judge} & & & & & & \\
Completeness & 0.22 & 0.50 & 0.59 & 0.57 & \textbf{0.62} & 0.53 \\
Visual Enrichment & 0.49 & 0.42 & \textbf{0.57} & \textbf{0.57} & 0.50 & 0.49 \\ \midrule
\textbf{Judge Averages} & & & & & & \\
Completeness & 0.25 & 0.51 & 0.62 & 0.58 & \textbf{0.64} & 0.55 \\
Visual Enrichment & 0.55 & 0.51 & 0.61 & \textbf{0.63} & 0.56 & 0.58 \\ \bottomrule
\end{tabular}
\caption{Comparative Performance of BWC Video Captioning Models Across Human and LLM-as-a-Judge Metrics}
\label{tab:results}
\end{table*}

As illustrated in Table \ref{tab:results}, the incorporation of a dedicated VQA stage consistently enhances performance across all metrics compared to the Direct VLM Summary baselines. Both QA-enhanced pipelines utilizing closed-source question generation (QG) models—Gemini 2.5 Flash and DeepSeek V3.2—demonstrate superior factual accuracy, completeness, and visual enrichment over their single-pass counterparts.

Our fine-tuned Qwen-based models generally exceed the performance of the VLM summary baselines and remain highly competitive with the closed-source QA pipelines. This suggests that structured question-answering effectively bridges the gap between mid-sized open-source models and flagship foundation models for domain-specific tasks.

In particular, the highest Human Factual Accuracy (0.91) was achieved by the Qwen 3 4B GRPO QA pipeline, while the strongest Completeness score (0.65) was attained by the Qwen 3 30B SFT QA variant. The closed-source QA pipelines scored highest in Visual Enrichment, with the DeepSeek-integrated model reaching a peak score of 0.68.

In comparing results using the QA pipeline and those that did not, the QA models showed improvement in performance over the baseline models, scoring between 0.04-0.13 points better in Completeness and 0.03-0.09 points better in Visual Enrichment. All of the QA pipelines also produced significantly denser captions than the baseline.

When comparing the results from the question generator using Foundation model pipelines (Gemini and Deepseek), and the Qwen3 finetuned models, it is clear that foundation models are better than the trained Qwen3 models; however, the performance between all of them did not differ by much, showing that these models are quite competitive and capable of putting up similar performance to models much larger than themselves.

\subsection{Ablation Study}

% ABLATION STUDY TABLE
% \begin{table}[h]
% \centering
% \caption{Ablation Study: Impact of Model Scale and Training Strategy on BodyCam-VQA Performance. (HFA: Human Factual Accuracy, Comp: Completeness, VE: Visual Enrichment).}
% \label{tab:ablation}
% \small
% \begin{tabular}{@{}lccc@{}}
% \toprule
% \textbf{Configuration} & \textbf{HFA $\uparrow$} & \textbf{Comp $\uparrow$} & \textbf{VE $\uparrow$} \\ \midrule
% \textit{Impact of Model Scale (SFT)} & & & \\
% Qwen3 4B SFT QA & 0.87 & 0.57 & 0.56 \\
% Qwen3 30B SFT QA & 0.86 & \textbf{0.64} & 0.56 \\ \midrule
% \textit{Impact of Training Strategy (4B)} & & & \\
% Qwen3 4B QA (Base) & 0.78 & 0.49 & 0.55 \\
% Qwen3 4B SFT QA & 0.87 & \textbf{0.57} & 0.56 \\
% Qwen3 4B GRPO QA & \textbf{0.91} & 0.55 & \textbf{0.58} \\ \bottomrule
% \end{tabular}
% \end{table}

\begin{table}[h]
\centering
\resizebox{\columnwidth}{!}{%
\begin{tabular}{@{}lccc@{}}
\toprule
\textbf{Configuration} & \textbf{HFA $\uparrow$} & \textbf{Comp $\uparrow$} & \textbf{VE $\uparrow$} \\ \midrule
\textit{Impact of Model Scale (SFT)} & & & \\
Qwen3 4B SFT QA & 0.87 & 0.57 & 0.56 \\
Qwen3 30B SFT QA & 0.86 & \textbf{0.64} & 0.56 \\ \midrule
\textit{Impact of Training Strategy (4B)} & & & \\
Qwen3 4B QA (Base) & 0.78 & 0.49 & 0.55 \\
Qwen3 4B SFT QA & 0.87 & \textbf{0.57} & 0.56 \\
Qwen3 4B GRPO QA & \textbf{0.91} & 0.55 & \textbf{0.58} \\ \bottomrule
\end{tabular}%
}
\caption{Ablation Study: Impact of Model Scale and Training Strategy on BodyCam-VQA Performance. (HFA: Human Factual Accuracy, Comp: Completeness, VE: Visual Enrichment).}
\label{tab:ablation}
\end{table}
To investigate model scaling and the various training methods employed, we conducted an ablation study using the Qwen3 model as the backbone. We examine the impact of SFT versus reinforcement learning (GRPO), as well as the effects of increasing parameter count.

As observed in Table \ref{tab:ablation}, the transition from the base Qwen model to ones that incorporate specialized QG training yields significant performance gains. The Base Qwen3 4B QG model, while utilizing our VQA architecture, achieved a human factual accuracy of only 0.78, which is lower than that of single-pass VLM Summary models. 

The most notable shift occurred with the implementation of the GRPO fine-tuned Qwen model. The GRPO variant was trained directly by a strong judge model, resulting in the highest factual accuracy (0.91) across all tested 4B models. For this test the reward function $\alpha$ and $\beta$ weights were each set to 0.5, to ensure an equal balance between structure and question quality. Interestingly, while GRPO slightly reduced the completeness score compared to SFT (0.55 vs. 0.57), it effectively minimized hallucinations, which is a critical requirement for forensic body-worn camera (BWC) documentation.

We further evaluated the effect of parameter scaling by comparing the 4B and 30B SFT variants. Increasing the model size to 30B parameters led to a substantial increase in Completeness (0.64), the highest in our study. This suggests that larger models are more capable of coming up with more specific and effective questions for investigation. However, this increase in information density did not translate to higher factual accuracy, which remained at 0.86. This finding reinforces our hypothesis that reinforcement learning (GRPO) is a more effective tool than raw scaling for ensuring the reliability of generated police reports.

\subsection{Case Study} 
% VIDEO 28 used here
% Other options: video279, 317, 58, are 1 for completness and visual enrichment too

To demonstrate the practical efficacy of the \textit{BodyCam-VQA} framework, we present a qualitative case study of a one-minute BWC video sequence. Generated captions alongside representative key frames are provided in Appendix~\ref{case_study} Figure~\ref{tab:case_study_frames}. This section serves as an illustrated example of the model's ability to capture high-stakes context in a real-world law enforcement scenario. The qualitative comparison of the generated captions reveals a clear progression in quality across the different model tiers, with our Fine-Tuned model performing the best. 

The selected clip features a response post-incident with the following ground-truth events:
\begin{itemize}
    \item \textbf{Initial Scene:} The officer observes fellow officers secure a scene at night in a residential driveway.
    \item \textbf{Analyze Surroundings:} The officer turns around to a residential street to gauge the surroundings.
    \item \textbf{Observation of Civilian:} The officer returns to the scene and observes the civilian on the ground.
    \item \textbf{Surveillance:} Fellow officers continue to monitor the scene, keeping a close-eye on the civilian. 
\end{itemize}

\noindent\textbf{Analysis:} 
Comparing the three captions, the Baseline Caption is noticeably shorter than the other two. When comparing the Baseline Caption to the QA Caption, the QA Caption provides substantially more detail, but it does so at the cost of accuracy. For instance, it incorrectly states that the scene takes place in an adjacent yard rather than in the present yard. It also misidentifies the siding material between two houses that were passed through by the bodycam officer. Additionally, it inaccurately captures details about the officers, such as how many are directing flashlights at the scene. The Baseline Caption also contains errors, such as misreporting the number of officers present and the number of officers using flashlights. 

When comparing both of these captions to the Fine-Tuned Caption, we observe clear improvement. The Fine-Tuned Caption avoids being overly brief like the Baseline Caption while also avoiding the excessive, speculative detail found in the QA Caption. Instead, it strikes a balance between conciseness and completeness. It captures important and accurate scene details such as the driveway setting, the presence of multiple flashlight beams, the civilian’s location and condition, and the fact that a weapon was recovered without introducing major hallucinations. 

The only error in the Fine-Tuned Caption is the incorrect attribution of flashlight use to the bodycam officer. This mistake also appears in the QA Caption and is likely due to the complexity of the scene (multiple officers; approximately five, including the bodycam officer) and multiple light sources create visual ambiguity. Given this, the error is understandable and relatively minor compared to the major inaccuracies found in the other captions. Overall, the Fine-Tuned Caption most effectively minimizes hallucinations while maintaining meaningful detail, making it the strongest and most balanced of them.

%% file: sections/5_conclusion.tex
\section{Conclusion and Future Work} 
% When using same model size, our method outperforms all baselines
%compared to closed source llms, models with too many parameters will be much stronger, however the performance is competitive
%Briefly restate what our pipeline is

BodyCam-VQA improves caption density for BWC footage by asking high-quality questions to the model, verifiably improving the model accuracy and enriching the output.
However, some issues remain in final caption generation, particularly the misattribution of physical attributes caused by model hallucination. Object misattribution was observed across all VLMs tested. The unique perspective of the camera and chaotic scenarios of BWC footage can lead to odd angles and clipping of objects in view, which leads to incorrect predictions for object recognition. Furthermore, since these models are given video and transcript data separately, determining the speaker at a given point in the video remains challenging. This problem is especially difficult since the video is taken from an officer's perspective, making it hard to determine whether a piece of dialogue came from the POV officer or another individual off-camera. Future work will focus on improving the model's ability to distinguish between individual people.

%% file: sections/6_limitations.tex
\section*{Limitations}
While BodyCam-VQA demonstrates consistent improvements in factual accuracy, completeness, and visual enrichment over baselines, we acknowledge several limitations of the current work:\\
$\bullet$ \textbf{\textit{Limited Temporal Scope.}} Our current pipeline and evaluation are constructed around 60-second video segments, a design choice made to keep frame sampling tractable and inference costs manageable within our computational budget. As a result, we do not evaluate the framework's ability to perform long-horizon temporal reasoning or to maintain coherent context across multiple linked segments from the same incident. Real-world BWC footage often spans several minutes to hours and may require linking evidence, entities, and events across segment boundaries to construct a complete investigative narrative. Extending BodyCam-VQA to support cross-segment memory and longer temporal windows is an important direction for future work.\\
$\bullet$ \textbf{\textit{Indirect Alignment in the Reward Signal.}} Our GRPO-based training relies in part on a model-based reward ($r_{judge}$) in which a large foundation model scores the relevance and investigative quality of generated questions. While this design allows us to scale supervision beyond hand-labeled question quality, it also means the policy is optimized toward what a judge model considers a high-quality question, rather than a signal directly derived from human forensic assessment. Since our human evaluation is applied only to the final generated captions and not to the intermediate question-generation reward, there remains a possibility that the judge's preferences diverge from those of human reviewers in edge cases, or that the policy could learn to exploit regularities in the judge's scoring rather than genuinely improving investigative value. We view incorporating direct human feedback into the reward loop, or periodically auditing judge-policy agreement, as a valuable next step to further ground this training signal.\\
$\bullet$ \textbf{\textit{Proof-of-Concept Evaluation Setting.}} Our evaluation is conducted using a curated ground-truth dataset and automated/human scoring of generated captions, which allows for controlled and reproducible comparison across model configurations. However, we have not yet evaluated BodyCam-VQA within an operational law enforcement workflow, such as direct comparison against officer-written reports, integration into report-drafting tools, or usability studies with investigators and legal reviewers. These deployment-oriented evaluations would help clarify the practical time savings, adoption barriers, and remaining accuracy gaps relevant to real-world use, and we consider this an important next stage beyond the current proof-of-concept study.

%% file: sections/7_ethics_statement.tex
\section*{Ethical and Societal Implications}
\addcontentsline{toc}{section}{Ethical and Societal Implications}
\phantomsection
\label{ethics}
The positive impact of our work lies in the enhancement of legal transparency, objective accountability, and the safeguarding of civil rights through AI-assisted forensic clarity. BWCs are intended to be impartial observers; however, the sheer volume and chaotic nature of this footage often lead to critical details being overlooked by human reviewers or hallucinated by standard VLMs. By structuring the captioning process through an adaptive VQA framework, our work directly contributes to a more equitable justice system, ensuring that latent nuances of suspect-officer interactions are accurately documented to protect both the public and law enforcement officers. We view the resulting increase in caption density and factual grounding, alongside a corresponding reduction in manual reporting burden, as a concrete step toward more reliable and equitable documentation of high-stakes encounters.

We use the publicly released COPA dataset for our experiments, which was made available for the purpose of civilian oversight, and we do not collect, process, or attempt to re-identify any footage or individuals beyond what is already disclosed in this source. Our use of this dataset and the overall design of our framework are consistent with the ACL Ethics Policy, and we do not foresee any additional ethical concerns beyond those already inherent to the publicly released source dataset.

\section*{Acknowledgment}
We gratefully acknowledge the support and collaboration from Worcester Polytechnic Institute and Axon. We also thank the reviewers and colleagues for their helpful comments and feedback.

%% file: sections/X_appendix.tex
\appendix
\clearpage 
\newcommand{\caseStudyTable}{%
  \small
  \begin{tabular}{cc}
  \toprule
  \multicolumn{2}{c}{\textbf{Key Frames}} \\
  \midrule
  \includegraphics[width=0.42\textwidth]{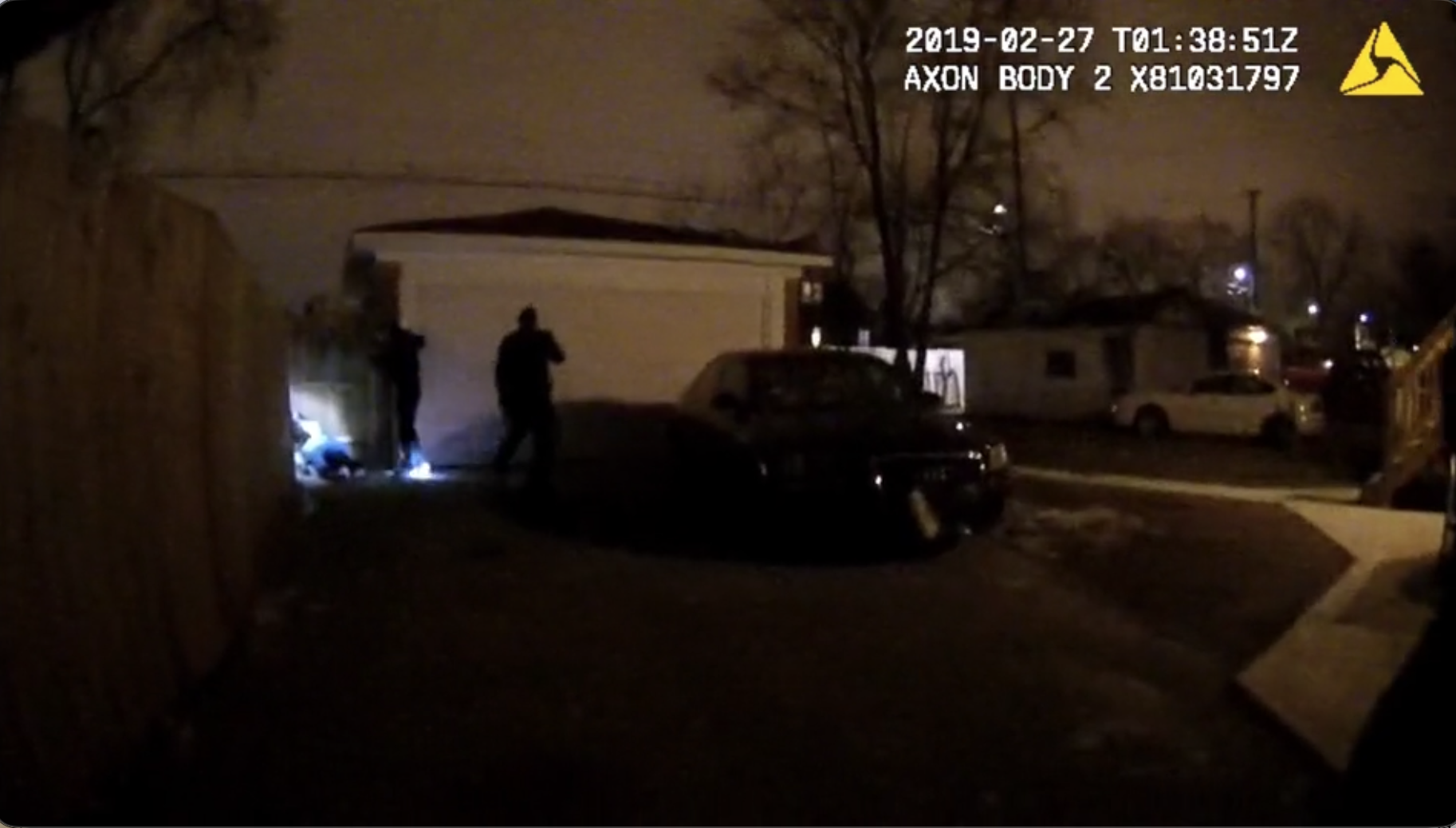} &
  \includegraphics[width=0.42\textwidth]{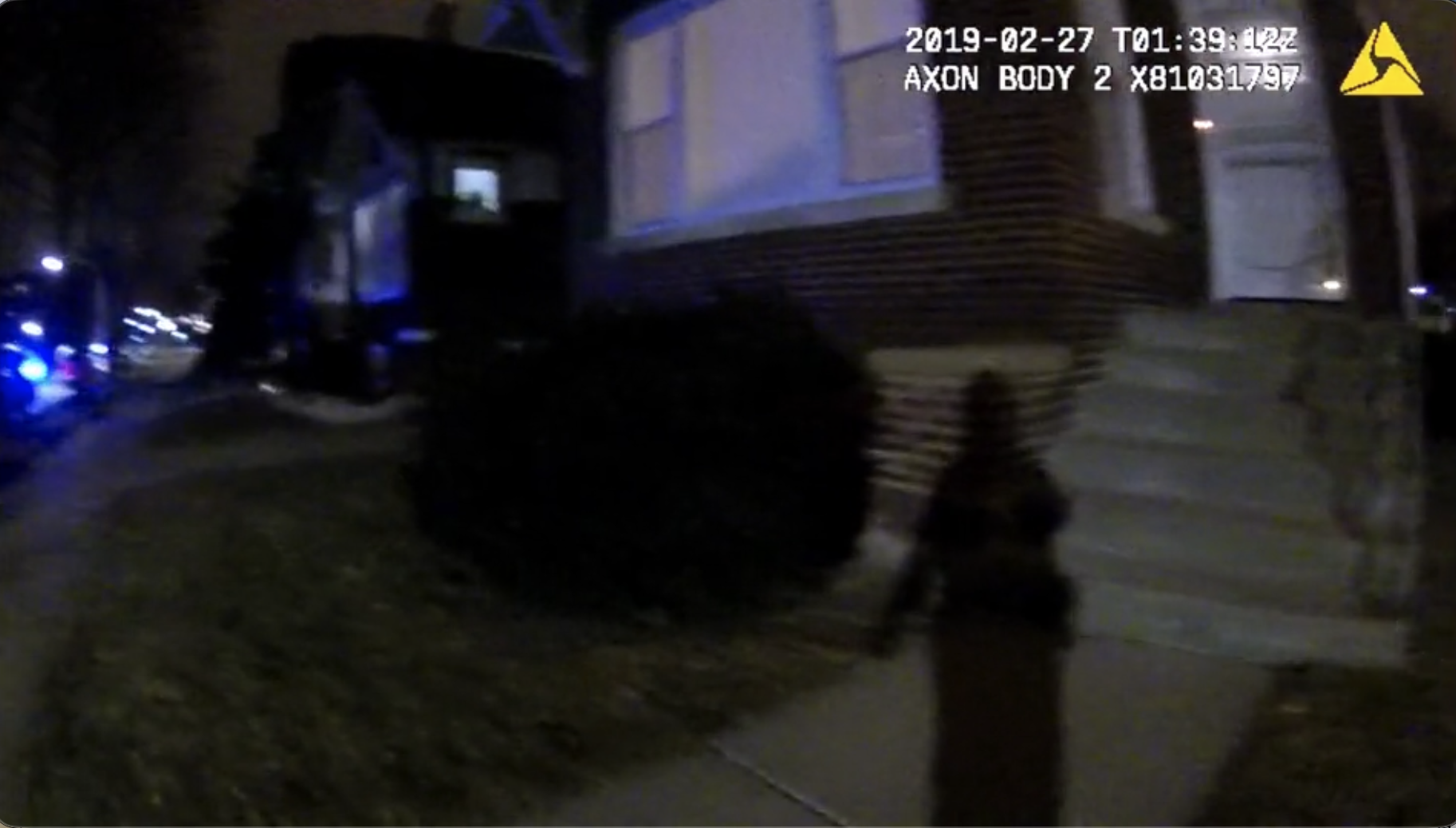} \\
  (a) Officers secure a scene at night in a residential driveway &
  (b) Emergency vehicles are present \\
  \midrule
  \includegraphics[width=0.42\textwidth]{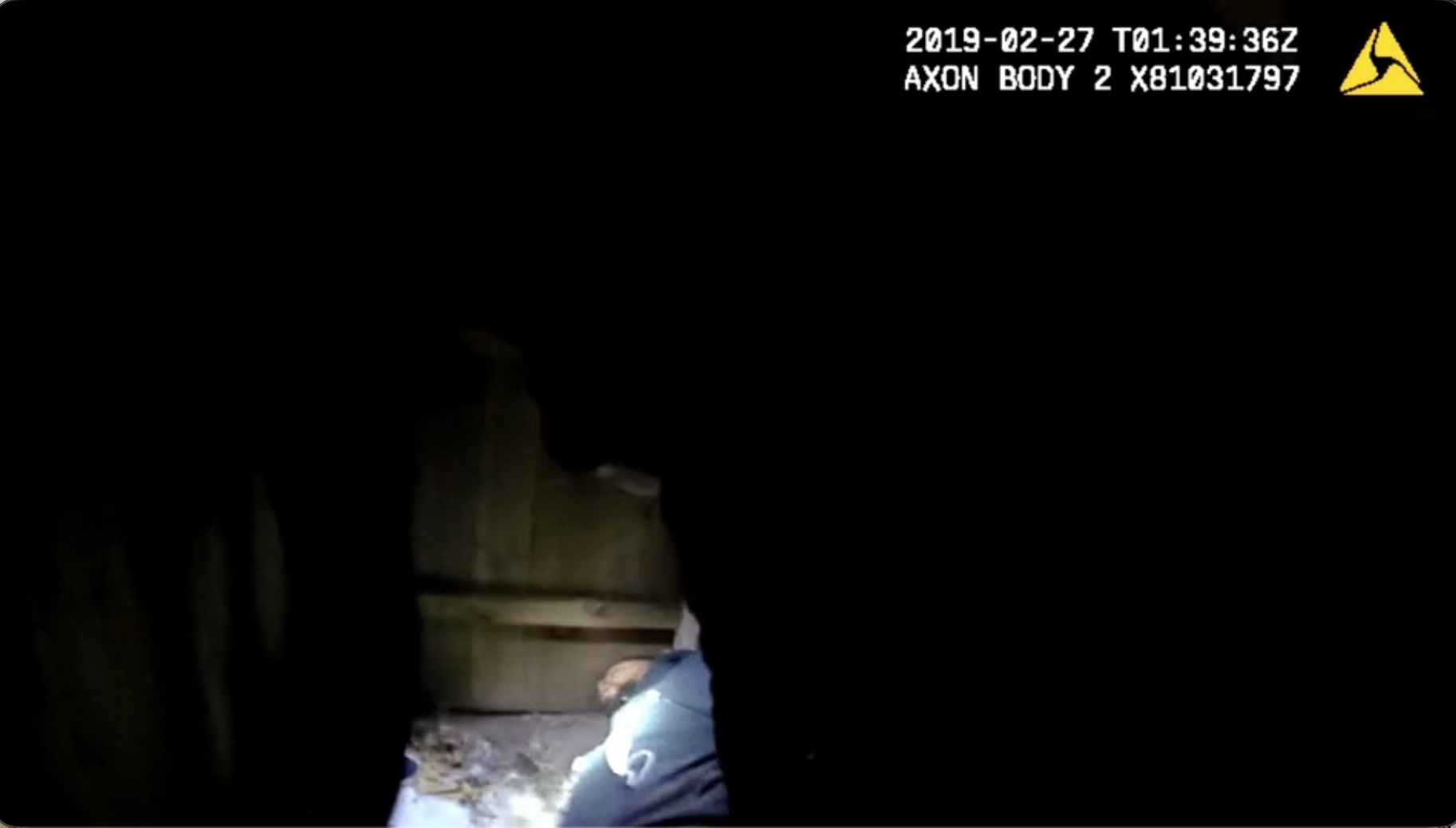} &
  \includegraphics[width=0.42\textwidth]{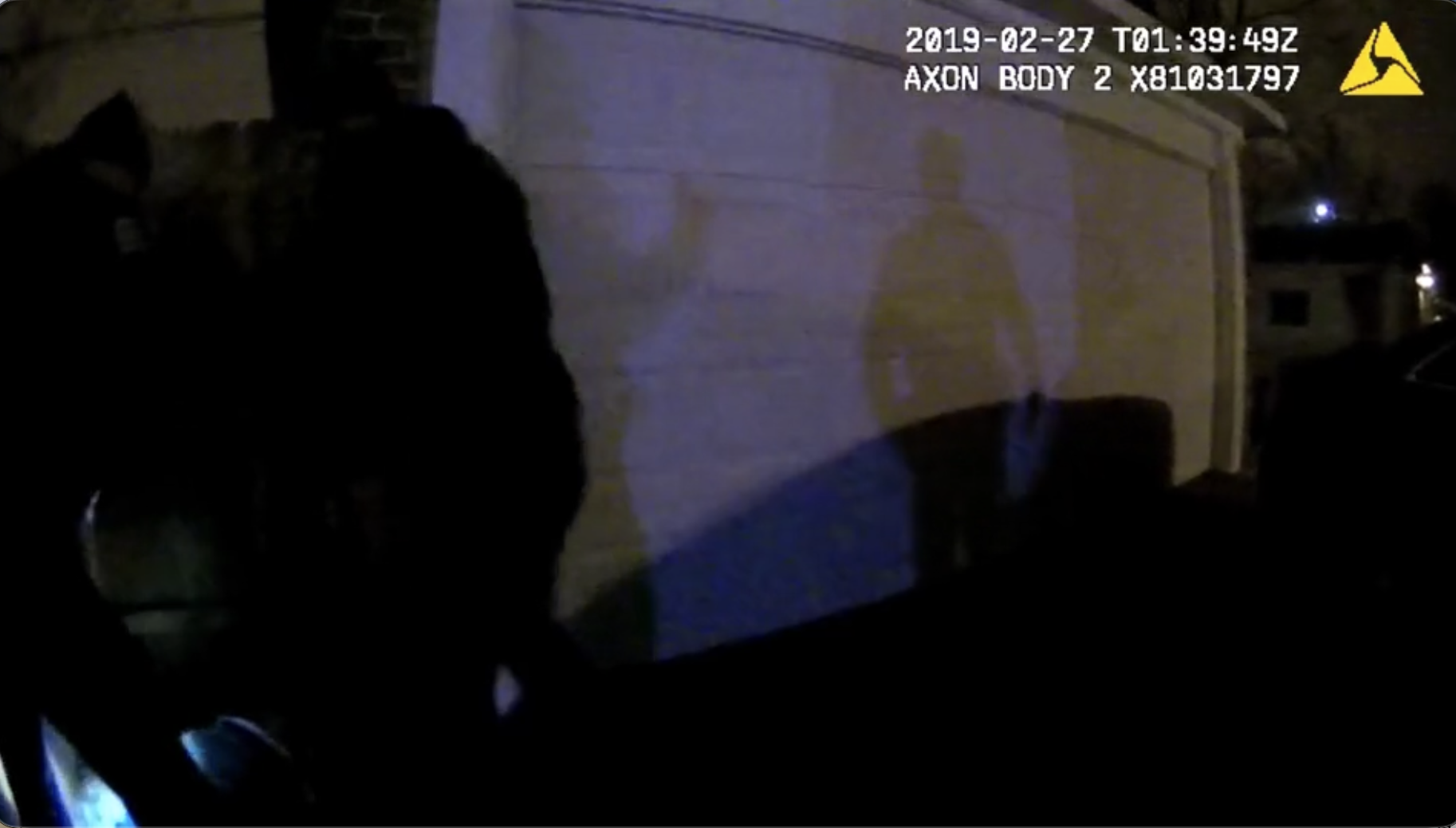} \\
  (c) Officers illuminate a subdued civilian on the ground &
  (d) Officers observe the subdued civilian \\
  \bottomrule
  \end{tabular}%
}

\twocolumn[
  \section{Case Study}
  \label{case_study}
  \centering
  
  \caseStudyTable

  \captionof{figure}{Representative Frames From Case Study. \textcolor{red}{Errors} in the generated captions below are highlighted in \textcolor{red}{Red}. \textbf{Notable/key information} is \textbf{bolded}.}
  \label{tab:case_study_frames}
  
  \vspace{2.5em} 
  \vspace{1.5em}
]

\noindent\textbf{Baseline Caption:} 
\textit{At night, police officers are in a \textbf{residential area} with single-family homes. A \textbf{dark sedan} is parked in a driveway \textbf{next to a white garage}. A person, identified as the \textbf{"offender," is on the ground next to the garage}. Police officers in dark uniforms, one wearing a black knit cap, attend to the person. \textcolor{red}{\textbf{One officer} uses a flashlight to illuminate the scene.} An officer states that a \textbf{weapon has been recovered} and requests \textbf{EMS}. The bodycam officer moves through a dark alleyway between houses, then walks on a sidewalk where police vehicles with \textbf{flashing blue lights are visible}. Sounds of a police siren are heard briefly, followed by a continuous phone ringing. The bodycam officer returns to the scene by the garage where \textcolor{red}{\textbf{three officers}} are surrounding the person on the ground. Officers repeatedly ask, "Hey, you okay?" and \textbf{instruct the individual to "Keep breathing."} An officer informs the person, "\textbf{EMS is on the way}, buddy," and continues to request EMS over the radio."}

\vspace{1em}

\noindent\textbf{QA Caption:} 
\textit{Nighttime envelops a residential property as an officer, wearing a dark uniform and body camera, approaches. This officer first focuses on a dark sedan parked in front of a white garage. \textcolor{red}{In the adjacent yard}, multiple officers, also in dark uniforms, attend to a \textbf{civilian lying on the ground}. An officer states, "offender is down on the ground in their yard," and another confirms a significant development, "\textbf{Weapon recovered}, let's get EMS back here," indicating a weapon was found on or near the civilian, though it remains unseen. The body camera officer rapidly navigates away from the immediate incident, moving down a narrow, dark alleyway \textbf{between two houses}, \textcolor{red} {one with dark siding and another lighter, where small, light-colored windows are briefly visible}. \textcolor{red}{Using a flashlight to cut through the darkness}, the officer quickly rounds a corner onto a street. There, active streetlights illuminate parked cars, including a \textbf{white pickup truck}, and the \textbf{flashing blue and red lights} of \textbf{emergency vehicles} are visible further down the road, indicating a strong police presence. The officer then passes the front of a residential house featuring a white front door with steps, brick siding, and an illuminated porch light, presumably moving to direct arriving medical personnel. Returning to the yard,\textbf{ \textcolor{red}{the body camera officer directs their flashlight to illuminate the scene where other officers assist the civilian.}}The civilian lies on their back on the dark, unpaved ground, which consists of dirt, gravel, and some debris next to the white garage. The \textbf{civilian wears dark long-sleeved clothing and dark pants.} While no major injuries or bloodstains are immediately visible, their breathing appears labored. The body camera officer asks, "Hey, you okay?" and instructs the civilian, "Keep breathing, dude," reassuring them, "\textbf{EMS is on the way}, buddy." Another officer states, "We're good, we're good," and the body camera user reiterates, "Everybody relax," and firmly instructs, "Squad, get EMS over here," emphasizing the urgent need for medical assistance. All visible doors and windows on the residential property appear closed, and \textbf{no other individuals are seen}, beyond the responding officers and the civilian, ensuring the scene's immediate security.}

\vspace{1em}

\noindent\textbf{Fine-Tuned GRPO Caption:} 
\textit{At night, \textbf{emergency vehicles with flashing blue lights line the street} of a residential area. A \textbf{dark-colored sedan} is parked \textbf{in a driveway}, nestled between a \textbf{white garage }and a \textbf{house}.\textbf{ Multiple flashlight beams}, along with \textbf{ambient light}, actively illuminate the scene, clearly detailing the sedan's wheels and body lines, the texture of the white garage door and wall, and the ground and driveway, including the tight space where a civilian is located. Officers, wearing dark uniforms, actively use their flashlights to pierce the darkness, \textbf{directing beams onto the scene}. One \textbf{officer recovers a weapon;} however, the bodycam officer does not handle it. A \textbf{civilian lies on the ground}, primarily \textbf{on their stomach or side,} in the tight space\textbf{ between the garage and a fence/wall}. The flashlight beams from various officers, along with a blue light, consistently illuminate the civilian's upper body and head. The \textbf{civilian appears subdued}, showing \textbf{no visible signs of active struggle, defensive postures, or rapid uncooperative movements,} and is breathing regularly, responding verbally with "Yeah" when an officer checks on their well-being. The bodycam officer initially approaches the scene where another \textbf{officer is already tending to the civilian} on the ground. After an officer states, "Weapon recovered." the bodycam officer immediately moves away, walking down a narrow path between houses and towards the street. After a brief pause near the street, the bodycam officer returns to the civilian's location near the garage. \textcolor{red}{The bodycam officer then gets closer to check on the civilian, shining their flashlight directly on them and instructing them to "keep breathing."} Officers reassure each other and the civilian, stating, "We're good," and "Everybody relax," as an officer reiterates the request for EMS to respond to their exact location."}